\documentclass{article} 
\usepackage{nips13submit_e,times}
\usepackage{hyperref}
\usepackage{url}
\usepackage{graphicx}
\usepackage{algorithm}
\usepackage{algorithmic}

\title{Spatial-Spectral Boosting Analysis for Stroke Patients' Motor Imagery EEG in Rehabilitation Training}

\author{
Hao Zhang, Liqing Zhang\\
MOE-Microsoft Key Laboratory for Intelligent Computing and Intelligent Systems\\
Department of Computer Science, Shanghai Jiao Tong University, China\\
\texttt{\{zh\_chaos, lqzhang\}@sjtu.edu.cn} \\
}

\nipsfinalcopy 

\begin{document}

\maketitle
\begin{abstract}
Current studies about motor imagery based rehabilitation training systems for stroke subjects lack an appropriate analytic method, which can achieve a considerable classification accuracy, at the same time detects gradual changes of imagery patterns during rehabilitation process and disinters potential mechanisms about motor function recovery. In this study, we propose an adaptive boosting algorithm based on the cortex plasticity and spectral band shifts. This approach models the usually predetermined spatial-spectral configurations in EEG study into variable preconditions, and introduces a new heuristic of stochastic gradient boost for training base learners under these preconditions. We compare our proposed algorithm with commonly used methods on datasets collected from 2 months' clinical experiments. The simulation results demonstrate the effectiveness of the method in detecting the variations of stroke patients' EEG patterns. By chronologically reorganizing the weight parameters of the learned additive model, we verify the spatial compensatory mechanism on impaired cortex and detect the changes of accentuation bands in spectral domain, which may contribute important prior knowledge for rehabilitation practice.
\end{abstract}

\section{Introduction}
\label{intro}
Brain Computer Interfaces (BCI) provides a communication system between human brain and external devices. The goal, transmitting control commands and feedbacks from brain cortex without utilizing human neural pathways, has been accomplished by signal recording and processing techniques based on BCI \cite{wolpaw2002brain}. Among assorts of brain diffused signals, electroencephalogram (EEG), which is recorded by noninvasive methods, is the most exploited brain signals in BCI studies.

Studies about motor imagery EEG have been widely undertook recently because of its discriminative property and inexpensive acquisitions. Thereinto, a novel application is combining BCI with clinical rehabilitation training therapies for strokes and related works have been conducted in \cite{buch2008think,guger2011state,meng2008bci}. In \cite{guger2011state}, an innovative system implementing BCI techniques in rehabilitation training is firstly introduced. In \cite{meng2008bci}, a simple paradigm inserting Functional Electrical Stimulation (FES) in BCI rehabilitation platform is proposed and Common Spatial Pattern (CSP) \cite{ramoser2000optimal} is employed as a feature extraction method for dealing with 2-class motor imagery. While, these studies concentrate more attentions on platform constructions but circumvent a difficult and important issue: EEG signals generated from impaired cortex differs from that of normal subjects in spatial and spectral patterns \cite{meng2008bci,liepert2000treatment,gourab2010changes,zhang2013gaussian}. Particularly, conventional feature extraction methods like CSP, Power Spectral Density (PSD) \cite{babiloni2000linear} or Phase Synchrony Rate (SR) \cite{NIPS2005_410} usually underperform when directly applied on EEG of strokes, only achieving a low level of classification accuracies and cannot detect gradual changes on frequency and locations over time or provide collaborative analytic interfaces. Moreover, they only give an ambiguous interpretation about the poor classification performance by introducing noise interferences but cannot provide a reasonable explanation from the point view of rehabilitation mechanism.
In our study, an integrated rehabilitation system, with active learning paradigm and multi-modal stimulations, is implemented for conducting a two months' clinical experiment \cite{zhang2013gaussian,Liu2013sigle}. In this paper, we focus on the offline analysis of stoke subject's 2-class motor imagery, specifically aiming at identifying the gradual changes of motor imagery EEG patterns of stroke patients in frequency and location during rehabilitation training and revealing potential mechanisms about stroke recovery, at the same time improving the classification performance by using spatial-spectral analytical methods.

Several studies about the compensatory mechanism (plasticity) of brain cortex \cite{meng2008bci,liepert2000treatment,liepert2000motor,Liu2013sigle} have suggested that before thorough recovery, the functions and responsibilities of the dysfunctional part of cortex will be partly undertook by its similar or neighboring parts. This inspires us that changing the channel configuration in rehabilitation studies is reasonable and effective when analyzing stroke subject' EEG. Instead of fixing the channel number as a stable count (16 or 64 \cite{meng2008bci,NIPS2012_0229}), we propose a dynamic channel selection strategy based on boosting \cite{freund1997decision,friedman2001elements} to verify the spatial migration caused by compensatory mechanism and to improve classification accuracies by adapting to this migration. The parameters of optimal channel groups in each iteration is recorded and then reformed, presenting an interesting spatial shift phenomenon on impaired cortex during rehabilitation training. Considering the frequency band, it is generally accepted that motor imagery accentuates EEG $\alpha$ and $\beta$ rhythms over sensorimotor cortices for normal subjects \cite{NIPS2005_410,pfurtscheller2001motor}. However, Gruab concludes in \cite{gourab2010changes} that motor processing in the brain is altered after chronic Spinal Cord Injury (SCI). Shahid reveals the $\mu$ and $\beta$ rhythm modulations in motor imagery related post-stroke EEG \cite{shahid2010mu}. A similar phenomenon is observed during our analysis, that is, the most significant frequency bands for discriminating 2 kinds of motor imagery have deviated from regular $\alpha$ and $\beta$ rhythms in different phases of the rehabilitation training \cite{Liang2013frequency}. Accordingly, an adaptive boosting technique dealing with frequency bands is proposed in this study for detecting the most informative spectral interval. By tracking the optimal bands automatically selected by boosting along with the training period, we discover an expanding tendency on spectral bands.
Moreover, spatial channel boosting (SB) and frequency band boosting (FB) are combined and then inserted as a complementary approach before extracting CSP features in the study. Comparative results are demonstrated, providing an evidence that spatial-frequency boosting (SFB) unearthes latent spatial-spectral information, which are often ignored during EEG feature extraction but indeed beneficial for discriminating stroke subjects' 2-class motor imagery.

\textbf{Difference with Existing Work} Various channel selection methods have been proposed in literature. Lan used a simple channel selection strategy to reduce feature dimensionality \cite{lan2007chanselec}. Mahnaz proposed a robust sparse common spatial pattern (RSCSP) algorithm to optimalize EEG channel selection for stroke subjects' EEG and compared it with other channel selection methods for normal subjects' EEG \cite{schroder2005robust,sannelli2010optimal}. Different from previous works, our study uses an adaptive boosting algorithm to detect the most competitive channel set on each day's training EEG of patients, and we emphasize the gradual changes of everyday's "best" channel set over time and attempt to connect this change with the cortex plasticity feature.
As for band selection, Pregenzer pointed out that the frequency component selection should be individual for each subject and also for each electrode position by using a DSLVQ relevance analysis \cite{pregenzer1999frequency}. Differently, our work is inspired by the frequency modulation proposed in \cite{gourab2010changes} and depends on stroke subjects' EEG. We are desired to detect the most active band in a more subdivided band set and to observe the band's gradual mutation along with the whole rehabilitation period so as to reveal potential rehabilitation mechanisms in spectral domain. Besides, we integrate channel selection with frequency selection to construct a combination classifier for improving the classification performance of BCI based rehabilitation system.

\textbf{Contributions}
In brief, the novelties of this paper can be summarized as follows:
(1) We address a new problem about analyzing stroke subjects' motor imagery EEG and emphasize several differences between this problem and classical 2-class motor imagery classification tasks.
(2) We develop a multi-modal BCI-FES rehabilitation system and have conducted a two months' clinical experiment in hospital.
(3) We propose a spatial-spectral selection algorithm based on boosting, which improves the classification accuracy for most subjects participated in our experiments. By utilizing intermediate outcomes of this algorithm, we detect the gradual changes in spectral and spatial domain and reveal some interesting phenomenons about stroke recovery.
(4) Combining with related studies about stroke rehabilitation, we give an interpretation about our two month's training performance and provide a novel prior knowledge for rehabilitation engineer.

In the rest part of the paper, the experimental arrangement and data acquisition are briefly described in Section \ref{data}.  Details of the algorithm is then illuminated in Section \ref{problem}. Section \ref{res} displays comparison results among CSP, SB, FB and their combinations, reorganizes some intermediate outcomes produced during boosting analysis and discusses the interconnections between the quantitative results and rehabilitation mechanisms. Finally, we give a brief conclusion about our work in Section \ref{conclusion}.

\section{Rehabilitation Paradigm and Data Collection }
\label{data}
Seven stroke subjects from hospital participated in our study as the experiment group. A control group with another 3 patients, only receiving traditional clinical treatments \cite{zhang2013gaussian}, was implemented for assessing the effectiveness of our rehabilitation platform. The medical diagnosis data of all 10 patients are recorded for comparisons (see supplementary material appendix E). Besides, we invited 3 normal subjects into our EEG collections for providing a contrast between stroke and normal ones. Some essential information about participants are provided in the supplementary material appendix A.
Eight weeks' training was conducted for data collection. Subjects were required to participate into 3 days' training per week, with 8 sessions per day. In each session, subjects would finish 15 trials of 2-class (left and right) motor imagery tasks. Each trial would last for 4 seconds with a randomized visual cue provided on the facing screen, instructing the subject imagine left or right timely. The 19-channel g.USBamp amplifier was adopted in our experiments. Raw EEG signals were recorded with sample rate 256Hz, stored and converted into .mat file in $Time*Channel*Trial$ format.

This paper adopts a new paradigm of BCI for rehabilitation training, which is briefly described as follows: firstly we use online BCI system to identify patients' motor imagery and record raw EEG signals  \cite{zhao2009eeg,NIPS2011_0748}, and then FES stimulations will be delivered to the corresponding limb with other multi-neurofeedbacks. To guarantee subjects' enough concentration during the continual and tedious training period, some active training paradigm such as acoustical prompts and interactive tasks are implemented in the paradigm \cite{liu2013active}. The new paradigm attempts to reconstruct the motor sensory feedback loop by establishing the causal relations between motor imagery and the actual limb movement \cite{zhang2013gaussian,Liu2013sigle}. More details about system configurations, paradigm and experiment parameters are provided in supplementary material appendix B. Note the low Signal-to-Noise ratio (SNR) of EEG, common-used preprocessing techniques are applied:
(1) Ground and reference electrodes: the forehead and post-ear electrodes are selected as reference and ground to eliminate the baseline (see supplementary material appendix B).
(2) Filtering: Bandpass filter is necessary to filter raw EEG into motor imagery related bands. In our experiment, we pre-filter the raw EEG signal into 5-40 Hz.
(3) Detrend: a simple detrend technique is employed for baseline drift correction.

\section{Problem and Algorithm}
\label{problem}
\subsection{Problem Modeling}
\label{modeling}
Two issues are often pre-decided as default in EEG analysis without deliberations: (1) How many and which channels should we take for analysis? (2) Which frequency band should we filter raw EEG signals into before feature extraction?
A steady configuration usually lose effects due to the cortex plasticity and spectral modulations mentioned above when dealing with strokes' EEG. Therefore, an improved dynamic configuration is required in this case.

For each subject, denote $E_t$ as the EEG data set of $t^{th}$ day's training ($1\le t \le60$, 60 days) after preprocessing and $E_{ti}$ as the $i^{th}$ trial of $E_t$ ($i \le N$). In our experiment, total 12 channels of EEG (see supplementary material appendix B) are taken into use so that $E_{ti}$ is a $1024*12$ matrix with a label $y_{ti}$ representing the direction of this imagery segment is left ($y_{ti} = -1$) or right ($y_{ti} = +1$).
In summary, our goal for spatial-spectral selection could be generalized as one problem, that is, under a universe set of all possible pre-conditions $\mathcal{V}$ we aim to find a subset $\mathcal{W} \subset \mathcal{V}$ which produces a combination model $F_t$ by combing all sub-model learned under condition $W_k (W_k \in \mathcal{W})$ and minimize the classification error on each days' data set $E_t$ (For convenience, we omit the time $t$):
\begin{equation}
\label{eq1}
\mathcal{W}^{*} = \arg\min_{\mathcal{W}} \frac{1}{N} |\{E_i: F(E_i, \mathcal{W}) \ne y_{i}\}_{i=1}^N|
\end{equation}
In the following part of this section, we will firstly model 3 homogeneous problems in detail and then propose an adaptive boost algorithm to solve them.

\textbf{Spatial Channel Selection}
Denote the universe set of all channels as $\mathcal{C} = \{C5, C6, FC3, FC4, \\C3, C4, CP3, CP4, P3, P4, C1, C2\}$
, of which each element is an electrode channel (we give a explanation about why choose these 12 channels in supplementary material appendix B). Denote $S$ as a channel set so that $|S| \le 12$.
Further, note that CSP usually outputs an even-dimensional feature which is a linear combinations of several different channels. To prevent a duplicated feature, we must guarantee $|S| \ge 4$ because the preferred feature dimension in this work is 4 (see supplementary material appendix B).
For convenience, we use a $12 \times 1$ binary vector to represent $S$, with 1 indicates this corresponding channel in $\mathcal{C}$ is selected while 0 not. Obviously the number of possible kinds of $S$ is $2^{12}-\sum_{i=0}^3 C_{12}^i$ = 3797.
Consider our original goal, we want to learn an optimal model which maximizes the classification accuracy on everyday's training data set by combining base classifiers learned under different $S$, at the same time record all used $S$ over time $t$ for temporal analysis. Denote $\mathcal{S}$ as a subset of $2^{\mathcal{C}}$, including all different $S$ contributing for learning $F$, then we get:
\begin{equation}
\label{eq2}
F(x, \mathcal{S}) = \sum_{k, S_k \in \mathcal{S}} \alpha_k f_k(x; S_k)
\vspace{-3pt}
\end{equation}
where $f_k$ is $k^{th}$ submodel learned with channel set precondition $S_k$, $\alpha$ is combination parameter. Multiplying $S_k$ leads to a projection on channel set $S_k$, which is the so-called channel selection.

\textbf{Frequency Band Selection}
Spectra is not a discreet variable as spatial channels. For simplication, we enable only the integer points on the closed interval $G = [5,40]$ (Hz) which is expanded from default $\alpha$ and $\beta$ band \cite{shahid2010mu} along the spectral axis.
Denote $B$ as a sub-band we split from global band $G$ and $\mathcal{D}$ as a universe set including all possible sub-bands produced by splitting. Note that the splitting procedure is supervised under following constrains:
\begin{itemize}
\vspace{-5pt}
\item Cover: $\bigcup_{B \in \mathcal{D}} B = G$
\item Length: $\forall B = [l, h] \in \mathcal{D}, 5\le h-l \le35$
\item Overlap: $\forall B_{min} = [l, l+1] \subset G, \exists B_1, B_2 \in \mathcal{D}, B_{min} \subseteq B_1 \cap B_2$
\item Equal: $\forall B_{min} = [l, l+1] \subset G, |\{B:B_{min} \subset B ,B \in \mathcal{D}\}| = C$, where $C$ is a constant
\vspace{-5pt}
\end{itemize}
These constrains guarantee that the set $\mathcal{D}$, consisted of finite sub-bands, will not underrepresents the original continuous interval and each band in $\mathcal{D}$ has an appropriate length. Accordingly, a sliding window strategy is proposed to produce $\mathcal{D}$ (see supplement material Appendix C). In our study, we produce almost 50 sub-bands to generate $\mathcal{D}$.

For band selection, we aim at detecting an optimal band set $\mathcal{B}$ ($\mathcal{B} \subset \mathcal{D}$), which is consisted of all active sub-band $B$ and produces an optimal combination classifier $F$ on everyday's training data:
\begin{equation}
\label{eq3}
F(x, \mathcal{B}) = \sum_{k, B_k \in \mathcal{B}} \alpha_k f_k(x; B_k)
\vspace{-4pt}
\end{equation}
where $f_k$ is the $k^{th}$ sub-model learned under band filter condition $B_k$. In our simulation study, a bandpass filter is employed to filter the raw EEG into band $B_k$.

\textbf{Combination}
For combining channel selection with frequency selection (SFB), we denote a two-tuple $(S, B)$ as a space-spectral condition where $(S,B) \in 2^{\mathcal{C}} \times \mathcal{D}$ and denote $\mathcal{U}$ as a set of all contributed two-tuples. Note that filtering and spatial projection is \textbf{interchangeable}, the combination function $F$ can be easily transformed as:
\begin{equation}
\label{eq4}
F(x, \mathcal{U}) = \sum_{k, (S,F)_k \in \mathcal{U}} \alpha_k f_k(x; (S, B))
\vspace{-10pt}
\end{equation}

\subsection{Learning Algorithm}
An adaptive boosting algorithm, mainly containing two steps, is proposed for dealing this problem.

\textbf{Training step}
This step models the different preconditions proposed above into different base learners. For each precondition $v_k \in \mathcal{V} (\mathcal{V} \in \{2^{\mathcal{C}}, \mathcal{D}, 2^{\mathcal{C}} \times \mathcal{D}\})$, the EEG segment $E$ are processed under condition $v_k$ and organized as a training data set $T_k$. CSP is employed to extract features from
$T_k$ and then a SVM classifier $f_k(x; \gamma(v_k))$ is trained, where $\gamma$ is the model parameter determined by both $v_k$ and $T_k$. This step establishes a one-to-one relationship between precondition $v_k$ and its related learner $f_k$ so that Equation \ref{eq1} can be transformed as:
\begin{equation}
 \min_{\{\alpha, v\}_0^K} \sum_{i=1}^N L(y_i, \sum_{k=0}^K \alpha_k f_k(x_i; \gamma(v_k)))
\label{eq5}
\end{equation}
where $K$ is the number of base learners (iteration times) and $L$ is the loss function.

\textbf{Greedy Optimization Step}
Equation \ref{eq5} can be solved with a greedy approach \cite{friedman2002stochastic,friedman2001greedy}. Note that
\vspace{-5pt}
\begin{equation}
F(x, \alpha, \gamma, V) =  \sum_{k=0}^{K-1} \alpha_k f_k(x; \gamma(v_k)) + \alpha_K f_K(x; \gamma(v_K))
\vspace{-5pt}
\end{equation}
we can conclude a simple recursion formula: $F_k = F_{k-1}(x) + \alpha_k f_k(x; \gamma(v_k))$.
To estimate $f_k$ and $\alpha_k$, we presuppose that $F_{k-1}$ has been determined so we get:
\vspace{-5pt}
\begin{equation}
F_k = F_{k-1}(x) + \arg \min_{f} \sum_{i=1}^N L(y_i, F_{k-1}(x_i) + \alpha_k f_k(x_i; \gamma(v_k)))
\label{eq7}
\vspace{-5pt}
\end{equation}
A steepest gradient descent \cite{friedman2001greedy} is introduced to minimize Equation \ref{eq7}. Given the pseudo-residuals:
\vspace{-5pt}
\begin{equation}
r_{\pi(i) k} = -\nabla_F L(y_{\pi(i)},F(x_{\pi(i)})) = - [\frac{\partial L(y_{\pi(i)}, F(x_{\pi(i)}))}{\partial F(x_{\pi(i)})}]_{F(x_{\pi(i)}) = F_{k-1}(x_{\pi(i)})}
\label{eq9}
\vspace{-5pt}
\end{equation}
where $\{\pi(i)\}_{i=1}^{\hat{N}}$ is the first $\hat{N}$ members of a random permutation of $\{i\}_{i=1}^N$. Then,
a new set $\{(x_{\pi(i)},r_{\pi(i)k})\}_{i=1}^{\hat{N}}$, which implies a stochastically-partly best descent step direction, is generated and utilized to learn the model parameter $\gamma(v_k)$:
\vspace{-5pt}
\begin{equation}
\gamma_k = \arg \min_{\gamma,\rho} \sum_{i=1}^{\hat{N}} [r_{\pi(i)k} - \rho f(x_{\pi(i)}; \gamma_k(v_k))]
\label{eq10}
\vspace{-5pt}
\end{equation}
As we have mentioned before, an one-to-one mapping between $\gamma_k$ and $v_k$ has been established so that we can naturally determine $v_k$ when $\gamma_k$ is definite. Note that in Equation \ref{eq9} we use a random subset $\{\pi(i)\}_{i=1}^{\hat{N}}$, instead of the full training data $\{i\}_{i=1}^N$, to fit the $k^{th}$ base learner $f_k$. This stochastic gradient is firstly introduced in \cite{friedman2002stochastic} to incorporate randomness in the stagewise iteration for improving performances. Different from the original stochastic gradient which use a completely random strategy, in our study we use a "Resample" heuristic for generating stochastic sequences. During the iteration process, we maintain a self-adjusted training data pool $\mathcal{P}$ at background. In each iteration, we select $\{\pi(i)\}_{i=1}^{\hat{N}}$ from $\mathcal{P}$ instead of from the original training set $\{x_i,y_i\}_{i=1}^N$, as Algorithm \ref{algo1} details. This strategy has been verified quite effective in our simulation studies because it not only conjoins randomness brought by stochastic gradient but also introduce a latent weighting mechanism for training samples that are false classified (see supplement material appendix D).
\vspace{-8pt}
\renewcommand{\algorithmicrequire}{\textbf{Input:}}
\renewcommand{\algorithmicensure}{\textbf{Output:}}
\begin{algorithm}[]
\caption{Resample Heuristic Algorithm for Stochastic Subset Selection }
\label{algo1}
\begin{algorithmic}[1]
\STATE Initialize the training data pool $\mathcal{P}_0 = T = \{x_i,y_i\}_{i=1}^N$;
\FOR{$k = 1$ to $K$}
\STATE Generate a random permutation $\{\pi(i)\}_{i=1}^{|\mathcal{P}_{k-1}|} = randperm(\{i\}_{i=1}^{|\mathcal{P}_{k-1}|})$;
\STATE Select the first $\hat{N}$ elements $\{\pi(i)\}_{i=1}^{\hat{N}}$ as $\{x_{\pi(i)}, y_{\pi(i)}\}_{i=1}^{\hat{N}}$ from $\mathcal{P}_0$;
\STATE Use $\{\pi(i)\}_{i=1}^{\hat{N}}$ to optimalize the new learner $f_k$ and its related parameters as in Algorithm \ref{algo2};
\STATE Use current local optimal classifier $F_k$ to split the original training set $T = \{x_i,y_i\}_{i=1}^N$ into two parts $T_{true} = \{x_i, y_i\}_{i:y_i = F_k(x_i)}$ and $T_{false} = \{x_i,y_i\}_{i:y_i \ne F_k(x_i)}$;\\

\hspace{-23pt}\textbf{Re-adjust the training data pool:\\}
        \FOR{each $(x_i,y_i) \in T_{false}$}
        \STATE Select out all $(x_i, y_i) \in P_{k-1}$ as $\{x_{i(m)}, y_{i(m)}\}_{m=1}^M$;
        \STATE Copy $\{x_{i(m)}, y_{i(m)}\}_{m=1}^M$ with $d (d\ge1)$ times so that we get total $(d+1)M$ duplicated samples;
        \STATE Return these $(d+1)M$ samples into $P_{k-1}$ and we get a new adjusted pool $P_{k}$;
        \ENDFOR
\ENDFOR
\end{algorithmic}
\end{algorithm}
\vspace{-8pt}
With $\gamma_k(v_k)$ , we can easily determine the combination coefficient $\alpha_k$ by solving:
\vspace{-6pt}
\begin{equation}
\alpha_k = \arg \min_{\alpha} \sum_{i=1}^N L(y_i, F_{k-1}(x_i) + \alpha f_k(x_i;\gamma_k(v_k)))
\label{eq11}
\vspace{-6pt}
\end{equation}
In summary, we give a simple framework of the whole process in pseudocode in Algorithm \ref{algo2} (leave out some details about resample heuristic, which has been detailed in Algorithm \ref{algo1}).

\textbf{Parameter Estimation}
Some remained problems about parameters determination is worth clarification. The iteration time $K$, which also determines the complexity of the final model $F$, is picked by using the early stopping strategy \cite{zhang2005boosting}. In our experiment, a combination model with almost 40 base learners is proved to have considerable generalization ability.
Consider $\hat{N}$, the size of the stochastic subset: if we decrease the ratio $ \hat{N}/N$, more randomness will be brought into the iteration, while, increasing this ratio provides more samples to train a more robust local base learner $f_k$. To choose an appropriate $\hat{N}$, we use model selection methods to search in a constrained range \cite{friedman2001elements}. In our simulation study we set $\hat{N}/N \approx 0.7$ and we have achieved a relatively satisfied performance and short training period.
In terms of $d$, the copies of incorrect-classified samples when adjusting $\mathcal{P}$, it is determined by the the local classification error $e = |T_{false}|/N$:
\vspace{-4pt}
\begin{equation}
\vspace{-5pt}
d = \max(1, \lfloor \frac{1-e}{e+\epsilon} \rfloor)
\end{equation}
where $\epsilon$ is an accommodation coefficient. Note that $e$ is always smaller than $0.5$ and will decrease during the iteration so that a larger penalty will be given on samples that are incorrect classified by stronger classifiers. This strategy warrants that the distribution of the samples in $\mathcal{P}$ will not change too much until $F$ has got a strong enough description ability about the training data.
As for the loss function $L$, we simply choose the squared error loss for calculation convenience. A future work is to be conducted for determining a better loss function.
\vspace{-8pt}
\renewcommand{\algorithmicrequire}{\textbf{Input:}}
\renewcommand{\algorithmicensure}{\textbf{Output:}}
\begin{algorithm}[]
\caption{The Framework of Spatial-Spectral Precondition Selection}
\label{algo2}
\begin{algorithmic}[1]
\REQUIRE
\boldmath
$\{x_i,y_i\}_{i=1}^N$: EEG training set of some day;
$L(y,x)$: The loss function;
$K$: The capacity of the optimal precondition set (number of base learners);
$\mathcal{V}$: A universal set including all possible preconditions;
\ENSURE
$F$: The optimal combination classifier;
$\{f_i\}_{i=1}^K$: The base learners;
$\{\alpha_i\}_{i=1}^K$: The weights of base learners;
$\{v_i\}_{i=1}^K$: The preconditions under which base learners are trained.
\unboldmath
\STATE Feed $\{x_i,y_i\}_{i=1}^N$ and $\mathcal{V}$ into CSP-SVM scheme to produce a family of base learners $\mathcal{F}$, so that a one-to-one mapping is established: $\mathcal{F} \leftrightarrow \mathcal{V}$;
\STATE Initialize $\mathcal{P}_0$, $F_0(x) = \arg \min_\alpha \sum_{i=1}^N L(y_i, \alpha)$;
\FOR{$k = 1$ to $K$}
\STATE Optimalize $f_k(x;\gamma(v_k))$ as described in \ref{eq10};
\STATE Optimalize $\alpha_k$ as described in \ref{eq11};
\STATE Update $\mathcal{P}_k$ as in Algorithm \ref{algo1} and $F_k(x) = F_{k-1}(x) + \alpha_k f_k(x;\gamma(v_k))$ ;
\ENDFOR
\STATE foreach $f_k(x;\gamma(v_k))$, use the mapping $\mathcal{F} \leftrightarrow \mathcal{V}$ to find its corresponded precondition $v_k$;
\RETURN $F$, $\{f_i\}_{i=1}^K$, $\{\alpha_i\}_{i=1}^K$, $\{v_i\}_{i=1}^K$;
\end{algorithmic}
\end{algorithm}

\vspace{-13pt}
\textbf{Complexity}
The computation cost of spatial-spectral selection algorithm mainly concentrates at the training step. In the training step, the spatial projection onto a channel set $S$ can be completed in $O(1)$ time. The spectral band filtering into a band $B$ can be finished with the time complexity of FFT at $O(n\log n)$. While, we are supposed to construct a universal precondition set $\mathcal{V}$ with at most 3797*50 possibilities in total, as mentioned in Section \ref{modeling}. Lastly, the time cost of extracting CSP features and training a SVM classifier is at worst case $O(DN^2)$, where $D$ is the dimension of feature. In our study, we choose $D = 4$, and the number of each day's training samples is approximate 2400. Taking consideration into the sparse characteristics of SVM and other kernel techniques, the cost of learning classifier can almost be ignored with regard to the precondition processing cost.
Fortunately, the spatial projection and spectral filtering, which is both independent with the optimalizatoin step, could be pre-processed and then stored in memory to avoid repeated computations in offline analysis. For future online usage, we can implement a multi-task scheduling to accelerate this procedure because projection or filtering at two different precondition is also independent with each other.
\vspace{-2pt}
\section{Result}
\label{res}
\begin{figure}[h]
\begin{center}
\includegraphics[width = 5in]{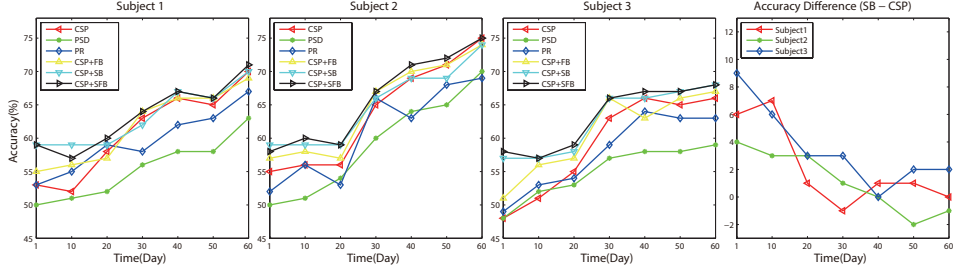}
\end{center}
\vspace{-10pt}
\caption{Testing accuracy of CSP, PSD, SR compared with utilizing our strategy SB, FB, SFB before extracting CSP features. Note that: (1) SB, FB and SFB achieves a higher classification accuracy than CSP in most cases; (2) A rising tendency is observed along with training time, which reflects that the rehabilitation training takes effects on these three subjects; (3) Subject 2 achieves a higher accuracy (mean value 0.7283) than subject 1 (0.6833) and 3 (0.6483) on the terminus of training, which is consistent with their medical assessments in supplementary material Appendix E.}
\label{accuracy}
\end{figure}

We evaluate our work mainly in three prospects: (1) The accuracy of our strategy compared with other commonly used methods about classifying 2-class motor imagery EEG. (2) Observations about the gradual changes of spatial-spectral preconditions over time. (3) Individual comparisons between the experiment group and normal subjects.
EEG data collected in each day (subject-independent) is split into 2 parts: The first 7 sessions for training and the last one for testing. 10-fold cross validation strategy is employed when training a combination classifier. Fig \ref{accuracy} gives the mean test accuracies of 7 days (1st, 10th, 20th, 30th, 40th, 50th, 60th day) achieved by several methods on subject 1,2,3 (see supplementary material appendix A and E). Note that PSD and SR features underperform on stroke subjects' EEG. A reasonable explanation is that PSD features excessively depends on the power spectrum and SR features is closely interrelated to changes of phase synchrony, while both of them are not so obvious and stable in patients' EEG over time compared with that of normal subjects.
\vspace{-10pt}
\vspace{-13pt}

It's worth emphasized that the accuracy improvements brought by SB is not identically distributed over time: the increment value appears larger at the beginning of training than at the end as shown in the 4th figure in Fig \ref{accuracy}. To explore the reasons, we exploit an advantage of boosting method: we have reserved all channel sets $S \in \mathcal{S}$ and sub-bands $B \in \mathcal{B}$ and their weights $\alpha$, which construct the classfication committee, to measure spatial-spectral changes.
We calculate a quantitative vector $L_t = \sum_{S_i \in \mathcal{S}} \alpha_i S_i $ on everyday's EEG to represent the \textbf{importance} of each channel in set $\mathcal{C}$. Similarly, the importance of each sub frequency band is calculated and then projected onto $[5,40]$.
\vspace{-5pt}
\begin{figure}[h]
\begin{center}
\includegraphics[width = 5in]{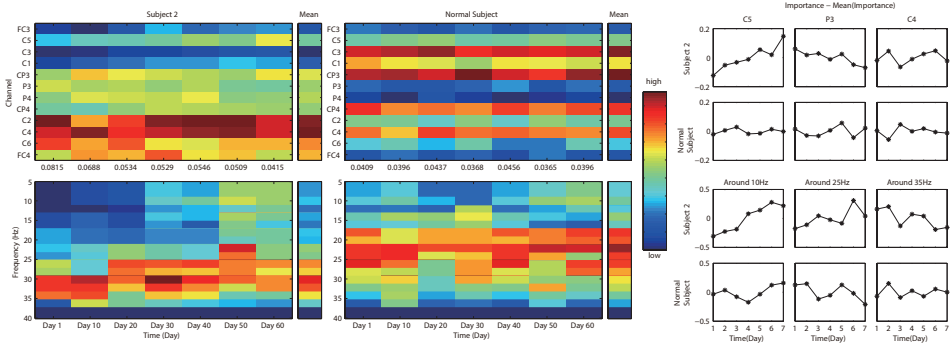}
\vspace{-12pt}
\end{center}
\caption{Left part: the importance of channels and bands, where red represents a higher importance and blue is lower. For obviating inner interferences, we measure the importance vector by applying SB and FB independently. The variances of normalized channel importance in each day are provided in the middle. Right part: the importance difference subtracting their average value in 7 days on channel C5, P3, C4 and sub-bands around 10Hz, 25Hz, 35Hz. Note that: (1) importance differences of (C5) and lower frequency (10Hz, 25Hz) gradually increase from negative to positive while that of the contralateral area holds; (2) The difference of P3 presents a decreasing tendency. (3) In most cases, the importance differences of normal subjects in both spatial and spectral domain maintains at a constant level.}
\label{importance}
\vspace{-12pt}
\end{figure}

Fig \ref{importance} illustrates 7 days' importance of channel and band for patient 2 together with a normal subject. Obviously we find that channels located at the patient's left cortex take a fewer effects than that of the right part, which is mainly caused by the hemiplegia on right limbs. The importance difference of left and right channels presents an abridge trend, that is, the importance of C3 and neighboring channels increases over time while importance of right channels decreases (in most cases), which indicates a rehabilitation sign about the impaired area. In more detail, the channel P3, which locates at the left part of cortex but may not been strongly impacted by stroke, have a considerable importance at the start of training. But different from other left channels, the importance of P3 decrease slightly over time. This special observation reminds us of the compensatory mechanism: P3 may initially takes parts of responsibilities of its neighboring area, which has been severely undermined by stroke, for motor imagery while this responsibility is slowly given back during recovery. Moreover, the variance of channel importance on each day is given at the middle of Fig \ref{importance}. For patient's EEG, variance at later period is much smaller than at the previous, while, the variance maintains at a relatively constant (low) level for normal subject. We conclude that a larger variance, which indicates a polarized distribution of channel weights, may lead a better performance of SB because futile channels will be eliminated by channel selection. This also supplies a justifiable explanation about the nonidentical-distributed increments of SB in Fig \ref{accuracy}: SB takes less effects in the later period because spatial patterns such as the distribution of each channel's weight appear more familiar with normal patterns.

\begin{figure}[h]
\begin{center}
\includegraphics[width = 5in]{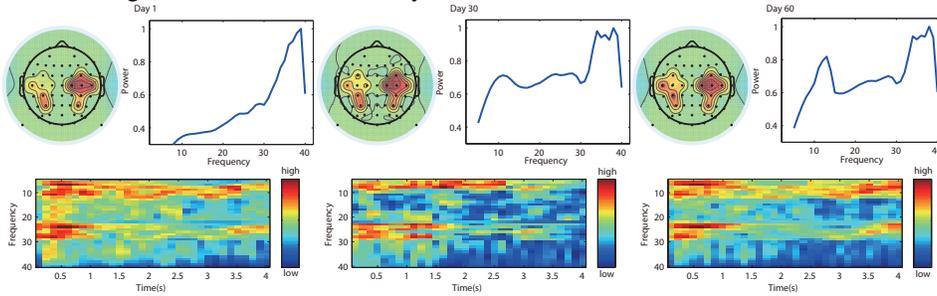}
\vspace{-15pt}
\end{center}
\caption{Detailed neural information, including spatial weights projection on scalp map, spectral-power curves and time-frequency characteristics, of subject 2 on day 1, 30 and 60. Note that: (1) Spatial weights of locations on impaired cortex gradually increase from margin to center. But the weight of C3 has very few increments compared to its neighbors. Similar changes could be also observed in Fig \ref{importance}. This phenomenon is exclusive on subject 2. Taking consideration into the highest accuracies achieved by subject 2, we sunrise that compensatory mechanism may have taken great effects on finishing right motor imagery for subject 2. (2) Power density \cite{babiloni2000linear} appears very large only on high frequency initially. But the power on $[8,30]$ Hz has increased slightly in the 60th day after training; (3) At the beginning of the training, ERD appears obviously in $[25,35]$ in patients' 4 seconds motor imagery in average. Along with the rehabilitation, ERD on low frequency band becomes more apparent. This is consistent with the band importance changes provided in Fig \ref{importance}.}
\label{combine}
\end{figure}

As for spectral bands, compared with normal subject, a higher band $[25,35]$ dominates at patient' motor imagery as shown in Fig \ref{importance}. But the importance of high frequency decentralizes, distributing to lower bands partly over time.
The slow migration on bands is exclusive on strokes' EEG when we conduct more observations on other healthy subjects. This dynamic band accentuation implies the frequency modulation mechanism during rehabilitation, and the increasing importance of $\alpha$ and $\beta$ rhythms indicates the recovery of related motor functions \cite{kirmizi2006comparative,kisley2006gamma}.
Essentially, to some extent, CSP features reflect the power changes generated by Event Related Desynchronization (ERD) on spectral band during motor imagery \cite{ramoser2000optimal,Liang2013frequency}.
The changes of accentuation bands in different training stages indicate that ERD in patients' motor imagery was broadened from high frequency bands to lower ones during rehabilitation, as Fig \ref{combine} illustrates. Therefore, FB provides a pathway to pick out the most active bands in different phase of training, and feed the weighted bands into CSP so that the band modulations could be detected and tracked by CSP, leading the improvements of discrimination ability of CSP features.
Compared with spatial changes, band selection shifts more slowly towards normal patterns over time. Moreover, the best performance achieved by SFB in Fig \ref{accuracy} evidence that SB and FB extract extra effective information in strokes' EEG, and they complement each other, both contributing to the classification ability of CSP features.
\vspace{-10pt}
\vspace{-10pt}

\vspace{-5pt}
\section{Conclusion}
\label{conclusion}
The proposed algorithm in this paper is the first work attempting to model the channel and frequency configuration as preconditions before learning base learners and then utilize an adaptive boosting strategy to construct an improved additive model. Similar with boosting, the algorithm produces a set consisting of the most contributed channel groups and frequency bands on each days' training EEG data. We compare classification accuracies of our algorithm with CSP, PSD, PR and the result demonstrates its competiveness. Furthermore, through tracking gradual changes of spatial and spectral preconditions selected by the algorithm along with the training time, we verified the compensatory mechanism on stroke cortex and observe a frequency expanding tendency over time. The discoveries are compared with chance. We believe the attempts in this paper provide a significant prior knowledge about stroke subjects' imagery pattern for future rehabilitation engineer.

\small
\bibliography{nips2013}
\bibliographystyle{unsrt}
\end{document}